\documentclass[aps,twocolumn,showpacs, tikz]{revtex4-2}
\usepackage{graphicx}
\usepackage{amsmath}
\usepackage{amssymb}
\usepackage{color}
\usepackage{float}
\usepackage{dcolumn}

\usepackage{tikz}
\usepackage{listofitems}
\tikzset{>=latex} 
\usepackage{xcolor}
\colorlet{myred}{red!80!black}
\colorlet{myblue}{blue!80!black}
\colorlet{mygreen}{green!60!black}
\colorlet{myorange}{orange!70!red!60!black}
\colorlet{mydarkred}{red!30!black}
\colorlet{mydarkblue}{blue!40!black}
\colorlet{mydarkgreen}{green!30!black}
\tikzstyle{node}=[thick,circle,draw=myblue,minimum size=22,inner sep=0.5,outer sep=0.6]
\tikzstyle{node in}=[node,green!20!black,draw=mygreen!30!black,fill=mygreen!25]
\tikzstyle{node hidden}=[node,blue!20!black,draw=myblue!30!black,fill=myblue!20]
\tikzstyle{node convol}=[node,orange!20!black,draw=myorange!30!black,fill=myorange!20]
\tikzstyle{node out}=[node,red!20!black,draw=myred!30!black,fill=myred!20]
\tikzstyle{connect}=[thick,mydarkblue] 
\tikzstyle{connect arrow}=[-{Latex[length=4,width=3.5]},thick,mydarkblue,shorten <=0.5,shorten >=1]

\begin{document}
\title{On the weight dynamics of learning networks}
\author{Nahal Sharafi}
\author{Christoph Martin}
\author{Sarah Hallerberg}
\affiliation{Hamburg University of Applied Sciences, Berliner Tor 21, 20099 Hamburg, Germany}
\date{\today}
\begin{abstract}

Neural networks have become a widely adopted tool for tackling a variety of problems in machine learning and artificial intelligence.
In this contribution we use the mathematical framework of local stability analysis to gain a deeper understanding of the learning dynamics of feed forward neural networks.
Therefore, we derive equations for the tangent operator of the learning dynamics of three-layer networks learning regression tasks. The results are valid for an arbitrary numbers of nodes and arbitrary choices of activation functions.
Applying the results to a network learning a regression task, we investigate numerically, how stability indicators relate to the final training-loss. 
Although the specific results vary with different choices of initial conditions and activation functions, we demonstrate that it is possible to predict the final training loss, by monitoring finite-time Lyapunov exponents or covariant Lyapunov vectors during the training process.
\end{abstract}
\maketitle
%
\section{Introduction}
Feed-forward networks trained via back-propagation are a frequently used tool in many applications 
\cite{ zhang2014fruit, talaat2020load,  ozanich2020feedforward, ketkar2021feed, ghani2021randomized, li2022intelligent, xie2022enhancing, ghani2023classification}.
Although the back-propagation process is well described as an algorithm, some aspects of its learning success still need to be investigated in detail.
Recent studies suggest to investigate learning networks from a  statistical physics point of view 
\cite{warner1996understanding, haber2017stable, kornblith2019similarity,  cocco2018statistical, bahri2020statistical, huang2021statistical, abbas2021power}.
Additionally, several contributions have also shown that analyzing neural networks as dynamical systems can effectively contribute to understanding the underlying processes and mechanisms of learning. \cite{weinan2017proposal, li2017maximum, liu2019deep,  engelken2023lyapunov, vogt2022lyapunov, chang2019antisymmetricrnn, li2022deep, tanaka2021dynamical}. 
Different studies have also investigated the stability properties of artificial neural networks \cite{gelenbe1990stability, fang1996stability, zhang2014comprehensive, saxeExactSolutionsNonlinear2014, drgonaSpectralAnalysisStability2020}.
In contrast to these approaches, investigating mostly the dynamics of network nodes, the focus of this study is the dynamics of the network's weights and biases during the training process.
Building on previous results for a linear network~\cite{saxeExactSolutionsNonlinear2014}, we derive equations for the dynamics of weights in feed-forward networks for arbitrary choices of activation functions.
We then characterize the learning dynamics of the networks by computing indicators of local stability, namely Lyapunov exponents (LEs), finite time Lyapunov exponents (FTLEs) as well as covariant Lyapunov vectors (CLVs)\cite{wolfe, ginelli2013covariant, Pazo, Parlitz}.
We find that analyzing the dynamics of the learning process reveals the influence of common choices of weight initializations \cite{he2015delving} on the variety of training-dynamics.
We show that the stability of different directions of the phase space have major effects on the final outcome of the training process.
As a practical consequence of our findings, we demonstrate that the learning outcome of training-runs (i.e., the final loss), can be predicted on the basis of FTLEs.

This paper is organized as follows: In Sec.~\ref{mod} we introduce the network model we are going to work with and the dynamical equations of the weight matrices.
Through out Sec.~\ref{jac} we derive the Jacobian matrix of the dynamical system describing the weight dynamics.
In Sec.~\ref{numres} we train our network to solve a regression task using 2 different kinds  of activation functions and compute their respective Lyapunov exponents in Sec.~\ref{le}.
In Sec.\ref{pred} we use our findings to predict the final loss of a training process at early steps of training and discuss the implications of our findings in Sec.~\ref{con}.

%
\section{Network model and weight dynamics}
\label{mod}
\begin{figure}[]
  \centering
\begin{tikzpicture}[x=2.2cm,y=1.4cm]
  \node[node in] (N1-1) at (0,1) {$a_1^{(0)}$};
  \node[node in] (N1-2) at (0,0) {$a_n^{(0)}$};
   \draw[connect, dotted] (N1-1) -- (N1-2);
   
    \node[node hidden] (N2-1) at (1,1) {$a_1^{(1)}$};
    \node[node hidden] (N2-2) at (1,0) {$a_n^{(1)}$};
    \draw[connect, dotted] (N2-1) -- (N2-2);
    
    \node[node out] (N3-1) at (2,0.5) {$a_1^{(2)}$};
  
    \draw[connect] (N1-1) -- (N2-1);
    \draw[connect] (N1-1) -- (N2-2);
    \draw[connect] (N1-2) -- (N2-1);
    \draw[connect] (N1-2) -- (N2-2);

    \draw[connect] (N2-1) -- (N3-1);
    \draw[connect] (N2-2) -- (N3-1);
  
  \node[above=5,align=center,mygreen!60!black] at (N1-1.90) {input\\[-0.2em]layer};
  \node[above=2,align=center,myblue!60!black] at (N2-1.90) {hidden layer};
  \node[above=8,align=center,myred!60!black] at (N3-1.90) {output\\[-0.2em]layer};
\end{tikzpicture}
  \caption{Visualization of a three-layer network with arbitrarily many nodes in the hidden layer. The regression task we present to this network requires two input nodes and one output node.}
  \label{fig:nnarchitecture}
 \end{figure}
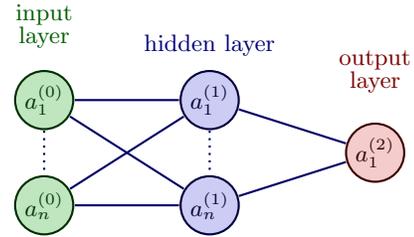
%
In this contribution we will consider a feed-forward network with three layers.
The network is trained via back-propagation, as it is common in classification and regression tasks.
Since we aim on formulating the weight dynamics analytically, we restrict this study to a network with just one hidden layer, but arbitrarily many nodes, as it is presented in Fig.~(\ref{fig:nnarchitecture}).
The number of the nodes in the input layer is $N_1$. The hidden layer has $N_2$ nodes and the output layer consists of $N_3$ nodes.

To describe the training process, we use a common textbook notation (see e.g., \cite{nielsen2015neural}) of input features, output, weight matrices and biases, i.e., the vector $\mathbf{x}$ represents the training input at each feed-forward pass, all training inputs are assembled within a matrix (or tensor) $\mathbf{X}$ and the vector $\mathbf{y}$ represents the target (also called ground truth).
Both $\mathbf{X}$ and $\mathbf{y}$ are given in the training-dataset within a a supervised learning scenario.

The feed-forward pass of input is described by the following equations (see e.g., \cite{nielsen2015neural}) for the state of the nodes in the hidden layer $\textbf{a}^{(1)}$ and $\textbf{a}^{(2)}$
\begin{eqnarray}
 \textbf{a}^{(1)} & = &\sigma(\mathbf{W}^{(21)}\mathbf{x} + \mathbf{b}^{(1)}), \\
 \textbf{a}^{(2)} & = & \mathbf{W}^{(32)}\textbf{a}^{(1)} +  \mathbf{b}^{(2)} =  \mathbf{W}^{(32)}\sigma(\textbf{d}) + \mathbf{b}^{(2)}. 
\end{eqnarray}   
Here, $\sigma$ denotes the activation function, the abbreviation
$\textbf{d} = \mathbf{W}^{(21)}\mathbf{x} + \mathbf{b}^{(1)}$ is introduced to shorten equations,
$\mathbf{b}^{(1)} \in {\rm I\!R}^{N_2}$ represents the bias of the hidden layer and
$\mathbf{b}^{(2)} \in {\rm I\!R}^{N_3}$ is the bias of the output layer; 
$\mathbf{W}^{(21)} \in {\rm I\!R}^{N_2 \times N_1}$ is the matrix of the weights between the first layer and the hidden layer and $\mathbf{W}^{(32)} \in {\rm I\!R}^{N_3 \times N_2}$ denotes the matrix of the weights between the hidden layer and the output layer.   

The cost function comparing the network output $\textbf{a}^{(2)}$ to $\textbf{y}$ is denoted by $C(\mathbf{y},\mathbf{a}^{(2)})$.
Using the gradient descent procedure (refered to as back-propagation in a machine learning context) in the continuous time limit \cite{backprop} we derive
the dynamics of the weights and biases,
\begin{eqnarray}
\dot{\mathbf{W}}^{(21)} &=& -(\sigma'(\mathbf{d}) \odot (\mathbf{W}^{(32)^T} \nabla C_{\mathbf{a}^{(2)}}))\mathbf{x}^T, \label{w21dot}\\
 \dot{\mathbf{W}}^{(32)} &=& -\nabla C_{\mathbf{a}^{(2)}}\sigma(\mathbf{d}^T),\label{w32dot} \\
\dot{\mathbf{b}}^{(1)} &=&  -\sigma'(\mathbf{d}) \odot (\mathbf{W}^{(32)^T}\nabla C_{\mathbf{a}^{(2)}}), \label{b1dot}\\
\dot{\mathbf{b}}^{(2)} &=& -\nabla C_{\mathbf{a}^{(2)}}, \label{b2dot}
\end{eqnarray}
where $\nabla C_{\mathbf{a}^{(2)}}$ is the gradient vector of the cost function with respect to $\mathbf{a}^{(2)}$ and  $\odot$ denotes the Hadamard product.
Note, that similar equations to describe the weight-dynamics in a three-layer network without activation function were employed in \cite{saxeExactSolutionsNonlinear2014}.
%
%
%
\section{Deriving the Jacobian of the training dynamics}
\label{jac}
All quantities needed for local stability analysis of a dynamical systems rely on the knowledge or the estimation of the Jacobian of the system.

Please note that we are referring to the Jacobian of the dynamical system described in Eqs.~(\ref{w21dot})-(\ref{b2dot}), which is {\sl not} identical to the matrix referred to as Jacobian within the backpropagation procedure.
In more detail, we are here computing partial derivatives of the temporal changes $\frac{d}{dt} W^{(...)}_{...}= \dot{W}^{(...)}_{...}$ and $\frac{d}{dt} b^{(...)} = \dot{b}^{(...)}$ in contrast to partial derivatives of weight values $W^{(...)}_{...}$ and bias values $\dot{b^{(...)}}$ required for backpropagation. 
As indicated above, the resulting Jacobian describes the tangent operator of the weight dynamics of a network with arbitrarily many nodes $N_1,N_2$ and $N_3$ in three layers.

\label{jacobian}
Assume the state of the system is a vector, consisting of consecutively arranged columns of $\mathbf{W}^{(21)}$, $\mathbf{W}^{(32)}$, $\mathbf{b}^{(1)}$ and  $\mathbf{b}^{(2)}$.
In the following the indices $i \in [0,N_1)$, $j \in [0,N_2)$ and $k \in [0,N_3)$ are used to describe positions in the matrices of weight and the vectors of biases. 
Using the mean-squared-error as a cost-function for a regression task, i.e., $ C(\mathbf{y},\mathbf{a}^{(2)}) = |\mathbf{y} - \mathbf{a}^{(2)}|^2$ one obtains the gradient as $\nabla C_{\mathbf{a}^{(2)}} =  2(\mathbf{a}^{(2)} - \mathbf{y})$. Taking the partial derivatives of Eq.(\ref{w21dot}) yields,
\begin{eqnarray}
\frac{\partial \dot{W}^{(21)}_{ji}}{\partial W^{(21)}_{lm}} &=& J_{iN_2+j, mN_2+l}, \\
&=& 2 x_i x_m (\delta_{jl}  \sigma''(d_{l})[W^{(32)^T}(y - a^{(2)})]_j\nonumber\\
&{\it }&-\sigma'(d_j)\sigma'(d_l)[W^{(32)^T}W^{(32)}]_{jl}),
\label{J1}
\end{eqnarray}
where $l \in [0,N_2)$, $m \in [0,N_1)$, $\delta$ is the Kronecker delta and $[\;]_i$ refers to the $i$th element of a vector and $[\;]_{i,j}$ to the element $i,j$  of a matrix.  
Similarly we obtain,
\begin{eqnarray}
\frac{\partial \dot{W}^{(21)}_{ji}}{\partial W^{(32)}_{nl}}   &=& J_{iN_2+j, N_1N_2 +  lN_3+n}, \nonumber \\
 &=&  2 x_i \sigma'(d_j)( \delta_{jl}  [y - a^{(2)}]_n -\sigma(d_{l})W^{(32)}_{nj}),  \quad \quad
\end{eqnarray}
with $n \in [0,N_3)$.
The derivatives with respect to the biases yield,
\begin{eqnarray}
 \frac{\partial \dot{W}^{(21)}_{ji}}{\partial b^{(1)}_l} &=& J_{iN_2+j, N_1N_2 + N_2N_3 + l}, \nonumber\\
 &=& 2 x_i(\delta_{jl} \sigma''(d_{j})[W^{(32)^T}(y -a^{(2)})]_j \nonumber \\
 &{\it }&- \sigma'(d_j)\sigma'(d_{l})[W^{(32)^T}W^{(32)}]_{jl}), \\
 \frac{\partial \dot{W}^{(21)}_{ji}}{\partial b^{(2)}_n} &=& J_{iN_2+j, N_1N_2 + N_2N_3 + N_2 + n},  \nonumber\\
 &=& -2  x_i \sigma'(d_j)W^{(32)}_{nj}.
\end{eqnarray}
Taking the partial derivatives of Eq.~(\ref{w32dot}) we arrive at the following elements of the Jacobian,
\begin{eqnarray}
 \frac{\partial \dot{W}^{(32)}_{kj}}{\partial W^{(21)}_{lm}} &=& J_{N_1N_2 + jN_3+k, mN_2 + l},  \nonumber\\
 &=&  2 \sigma'(d_l)x_m(\delta_{jl}[y - a^{(2)}]_k - \sigma(d_j)W^{(32)}_{kl}),\quad \quad\\
 \frac{\partial \dot{W}^{(32)}_{kj}}{\partial W^{(32)}_{nl}} &=& J_{N_1N_2 + jN_3+k, N_1N_2 + lN_3 + n}, \nonumber \\
 &=& -2 \delta_{kn} \sigma(d_j) \sigma(d_l),\\
 \frac{\partial \dot{W}^{(32)}_{kj}}{\partial b^{(1)}_l} &=& J_{N_1N_2 + jN_3k, N_1N_2 + N_2N_3 + l}, \nonumber \\
 &=& 2\sigma'(d_{l}) ( \delta_{jl}  (y - a^{(2)}]_k - \sigma(d_j)W^{(32)}_{kl}),\\
\frac{\partial \dot{W}^{(32)}_{kj}}{\partial b^{(2)}_n} &=& J_{N_1N_2 + jN_3+k, N_1N_2 + N_2N_3 + N_2 + n},  \nonumber \\
 &=& -2 \delta_{kn} \sigma(d_{j}).
\end{eqnarray}
%
Computing the partial derivatives of Eq.~(\ref{b1dot}) we obtain,
\begin{eqnarray}
 \frac{\partial \dot{b}_j^1}{\partial W^{(21)}_{lm}} &=& J_{N_1N_2 + N_2N_3 + j, mN_2 + l},\nonumber \\
 &=&   2 x_m (\delta_{jl}\sigma''(d_j)[W^{32^T}(y - a^{(2)})]j\nonumber\\
 &{\it }& -\sigma'(d_j)\sigma'(d_l)[W^{32^T}W^{(32)}]jl),\\
 \frac{\partial \dot{b}_j^{(1)}}{\partial W^{(32)}_{nl}} &=& J_{N_1N_2 + N_2N_3 + j, N_1N_2 + lN_3 + n}, \nonumber \\
 &=& 2 \sigma'(d_j) (\delta_{jl}[y - a^{(2)}]_n - \sigma(d_l) W^{(32)}_{nj}),\quad \quad\\
 \frac{\partial \dot{b}_j^1}{\partial b_l^{(1)}} &=& J_{N_1N_2 + N_2N_3 + j, N_1N_2 + N_2N_3 + l}, \nonumber\\
 &=&   2 (\delta_{jl}\sigma''(d_j) [W^{(32)^T}(y - a^{(2)}]_j \nonumber\\
 &{\it }&-\sigma'(d_j)\sigma'(d_l) [W^{(32)^T}W^{(32)}]jl),\\
 \frac{\partial \dot{b}_j^{(1)}}{\partial b_n^{(2)}} &=& J_{N_1N_2 + N_2N_3 + j, N_1N_2 + N_2N_3 + N_2 +  n}, \nonumber\\
 &=& -2 \sigma'(d_j) W^{(32)}_{nj}.
\end{eqnarray}
%
And similarily the partial derivatives of Eq.~(\ref{b2dot}) are,
\begin{eqnarray}
 \frac{\partial \dot{b}^{(2)}_k}{\partial W^{(21)}_{lm}} &=& J_{N_1N_2 + N_2N_3 + N_2 + k, mN_2 + l}, \nonumber\\
 &=& -2 \sigma'(d_l) W^{(32)}_{kl} x_m, \\
 \frac{\partial \dot{b}^{(2)}_k}{\partial W^{(32)}_{nl}} &=& J_{N_1N_2 + N_2N_3 + N_2 + k, N_1N_2 + lN_3 + n}, \nonumber \\
 &=& -2 \delta_{kn} \sigma(d_l),\\
 \frac{\partial \dot{b}^{(2)}_k}{\partial b^{(1)}_l} &=& J_{N_1N_2 + N_2N_3 + N_2 + k, N_1N_2 + N_2N_3 + l}, \nonumber\\
 &=& -2 \sigma'(d_l) W^{(32)}_{kl},\\
\frac{\partial \dot{b}^{(2)}_k}{\partial b^{(2)}_n} &=& J_{N_1N_2 + N_2N_3 + N_2 + k, N_1N_2 + N_2N_3 + N_2 + n}, \nonumber\\
 &=& -2 \delta_{kn}. \label{J16}
\end{eqnarray}
In addition to deriving the equations for the Jacobian rigorously, we tried to infer the entries of the Jacobian from data, as e.g., proposed in \cite{Martin2022estimating}.
However, the matrices of candidate functions which are an important step in the procedure \cite{Martin2022estimating}, were not sparse enough and consequently better results were achieved using weighted direct numerical derivatives.
For the shallow network studied in this contribution, the numerically estimated Jacobians compared well to the rigorous results, indicating that similar studies on deeper networks could also be undertaken employing numerically estimated Jacobians.
%
\section{Numerical results for a specific regression task}
\label{numres}
In the following, we test the capabilities of our framework by analyzing the training process of various networks while solving a regression task.
The training data for the non-linear regression problem consists of $n=1000$ observations of two features, $x_1$ and $x_2$, which are normally distributed with mean $\mu = 0$ and standard deviation $\sigma = 1$;
$x_1, x_2  \sim \mathcal{N}(0,1)$.
%
\begin{figure}[t!]
  \includegraphics[width=0.4\textwidth]{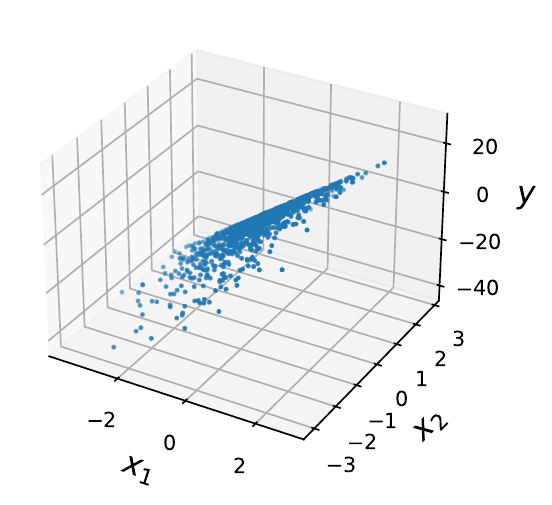}
 \caption{\label{fig:traindata3d} Visualization of the two-dimensional training data used in this paper. The quadratic relationship between the variables is discernible in the figure.}
\label{vis}
 \end{figure}
%
The target $y$ is a non-linear combination of $x_1$ and $x_2$: $y = 10 x_1 - 2 x{_2}^2$.
Features and target are visualized in Fig.~\ref{fig:traindata3d}.

In order to predict the target, we use a minimalistic version of the a three-layer network (see Fig.~\ref{fig:nnarchitecture}) with a hidden layer of two neurons, i.e.,~$N_2 = 2$, the input layer $N_1 = 2$ and $N_3=1$ neuron in the output layer.
These five neurons are connected by 6 weights. In the hidden layer and the output layer 3 biases are added, which are also subject to changes through gradient descent.
Consequently the changes of all trainable parameters during the training-process can be understood as a nine-dimensional dynamical system.
Additionally the dynamics is influenced by external ``perturbations'', (i.e., input data $\mathbf{X}$ and target $\mathbf{y}$) of the training-data-set, which enter the cost-function, driving the dynamics of weight-updates. 
Since the training-data is given and remains unaltered by the dynamical system of weight-changes, yet it influences every time-step (i.e., here weight update), one can question whether it is valid to interpret the training-data set as a stochastic part of the dynamics.
Following this idea, one could potentially  study, the influence of different training-data sets and machine learning tasks on the same network-structure.
In this contribution, however, we restrict ourselves to investigating the influences of weight-initialization and activation functions on the weight-dynamics, while using an identical task and training-data-set.
More precisely, we trained the network using either rectified linear unit (ReLU) (see e.g.,~\cite{fukushima1969visual}), Gaussian error linear unit (GELU) (see e.g.,~\cite{hendrycks2016gaussian}) or tanh as activation functions.
In this contribution we present the results with tanh and ReLU activations, since there was no qualitative difference between ReLU and GELU results for the specific machine learning task we studied.

Note that the exact solution to this regression task can not be reached with a network of this size.
Although we can reach quite small loss values we will not reach zero loss during trainings.
Hence we observe the dynamics of a system, presented with a task that is only partially solvable.

\begin{figure}[t!]
  \includegraphics[width=0.5\textwidth]{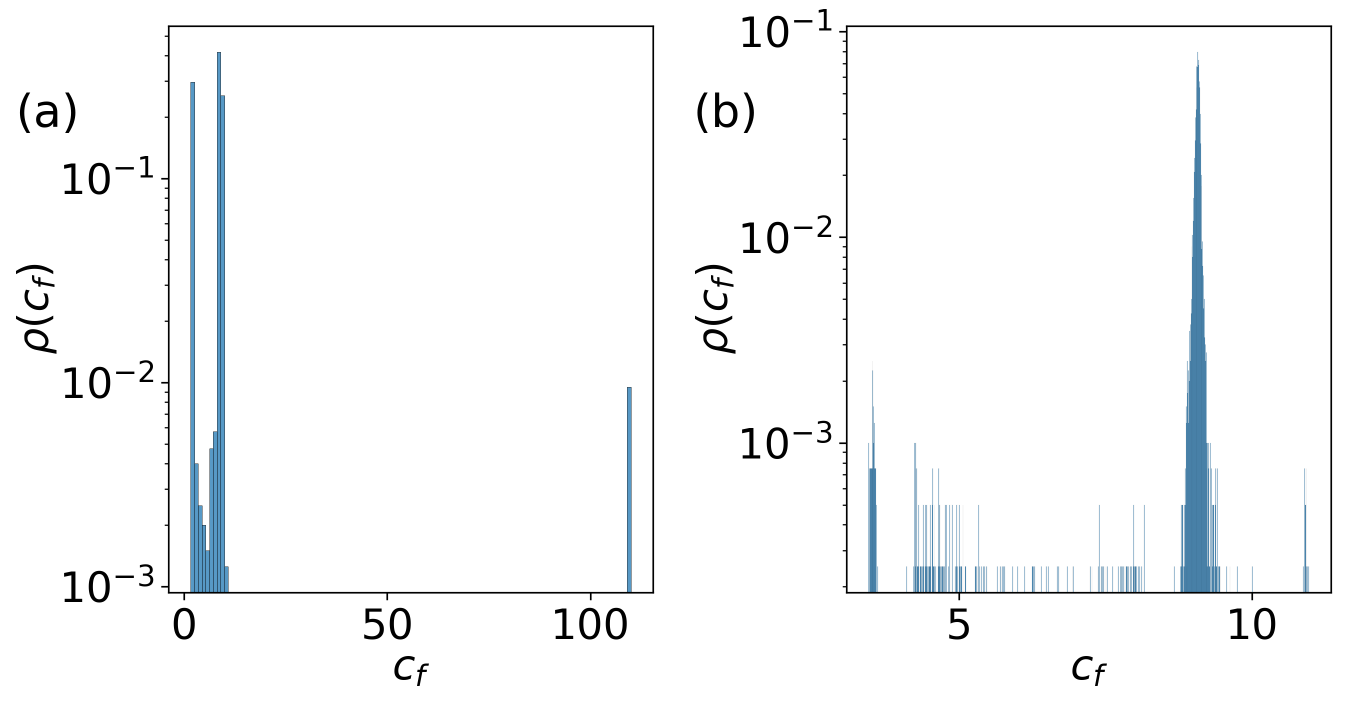}
 \caption{\label{2losshist} Distributions of final losses $c_f$ (values of the cost-function at the end of a training run) for 8000 realizations with He initializations exhibit distinct clusters. (a) Final loss values of 4000 network realizations using ReLU as activation. (b) Final loss values of 4000 network realizations using tanh as activation function.}
\end{figure}
 To investigate the dependence on the initial conditions, we compare two different methods of weight initializations and 4000 different training runs, each initialized with a different randomly drawn set of initial conditions.
 A common and highly recommended initialization method for nonlinear activation functions, is the Kaiming He initialization \cite{he2015delving,glorot2010understanding, kumar2017weight}.
 The He method is suitable for nonlinear activation functions as it avoids explosion or vanishing of the input due to exponential growth.
 This is achieved by restricting the standard deviation of random numbers used for weight initialization to $\sqrt{\frac{2}{N}}$, ($N$ being the number of neurons in the corresponding layer).
 In contrast to this, we also test an ad-hoc-choice of initial conditions drawn from a Gaussian distribution with standard deviation of $\sigma = 20$, which we will refer to as wide-range-initialization in the following. 

 For 8000 training runs initialized with He initialization and 8000 runs initialized with wide-range initialization we measure the final losses $c_f$ (i.e., values of the cost function at the end of the training) and compute indicators of local stability (finite-time Lyapunov exponents and covariant Lyapunov vectors).
 Half of these realizations of the network use ReLU as an activation function, the other half employs tanh. 
 \begin{figure}[t!]
  \includegraphics[width=0.5\textwidth]{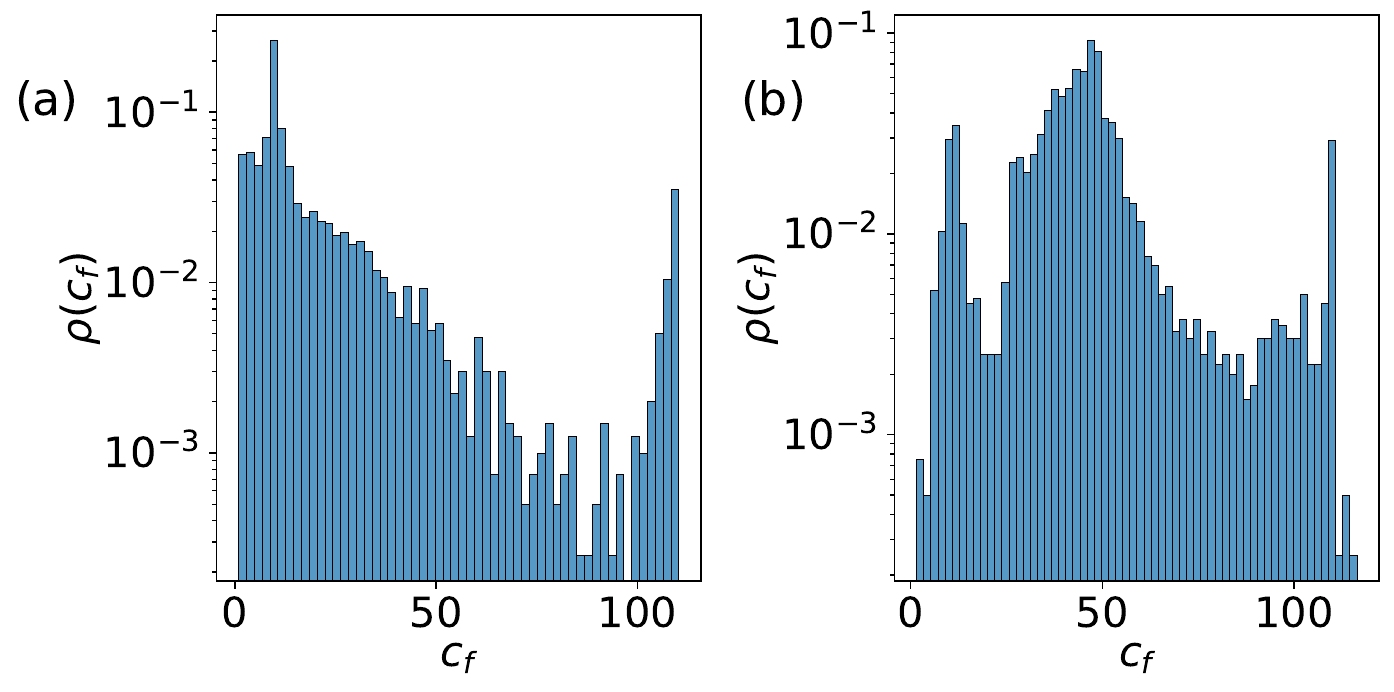}
  \caption{\label{2wlosshist} Wide range initialization of network weights creates a wider distribution of training results, allowing non-optimal training results to occur more frequently.
    Distributions of final losses $c_f$ (values of the cost-function at the end of a training run) of 8000 realizations with wide range random initializations, $\sigma = 20.$.}
\end{figure}
  As can be seen in Fig.~\ref{2losshist}(a), (combining He initialization and ReLU) the distribution $\rho(c_f)$ of the loss values suggest existence of three different attractors (i.e., local minima of the loss function).
 One corresponds to low values of the loss function, another to intermediate loss values and a third one to high loss values %
 of above 100.
 This high loss values are absent if the network is trained with a tanh as an activation function as it is shown in Fig.~\ref{2losshist}.(b).
However there is still a distinct peak in the distribution which corresponds to low loss values.  
As can bee seen in in Fig.~\ref{2wlosshist} starting with wide-range initialization, the distributions of loss values are flatter and wider compared to the He initialization and higher loss values occur even if tanh is used as activation function.
Although specific peaks of these distributions are recognizable, they are not as pronounced as in the case of He-initialization and values between peaks are observed frequently.
Interpreting peaks in loss-distribution as an indicator for different local minima of the loss-function one could suspect that wide range initialization prevents a fast convergence to these minima. In terms of dynamical systems theory the local minima correspond to attractors in the phase space.
One potential explanation for this could be that larger initial conditions are amplified in the network, which would in a deeper network cause numerical instabilities, but only delays the convergence to the next local minimum in a smaller network. Moreover since new peaks in the distribution of losses are discernible we can speculate that wider range of initial conditions makes basin of attraction of a larger number of attractors accessible.
\section{Characterizing the training process through indicators of local stability}
\label{le}
\begin{figure*}[t!!!]
 	\centering
 	\includegraphics[width=\textwidth]{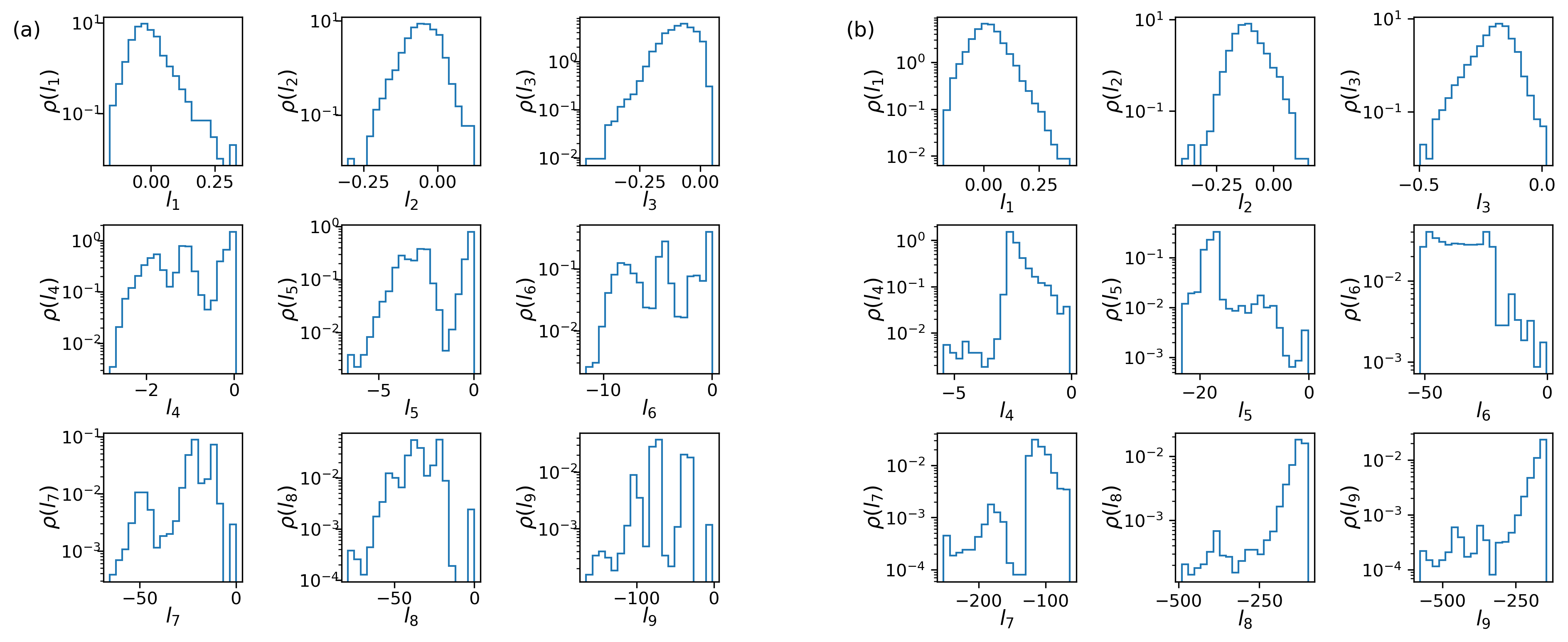}
 	\caption{Distributions of Lyapunov exponents computed for 8000 training runs of networks initialized with He initialization, amplifying input with (a) ReLU as activation function and (b) tanh as activation function.}
 	\label{lehist}
 \end{figure*}
\begin{figure*}[t!]
  	\centering
  	\includegraphics[width=\textwidth]{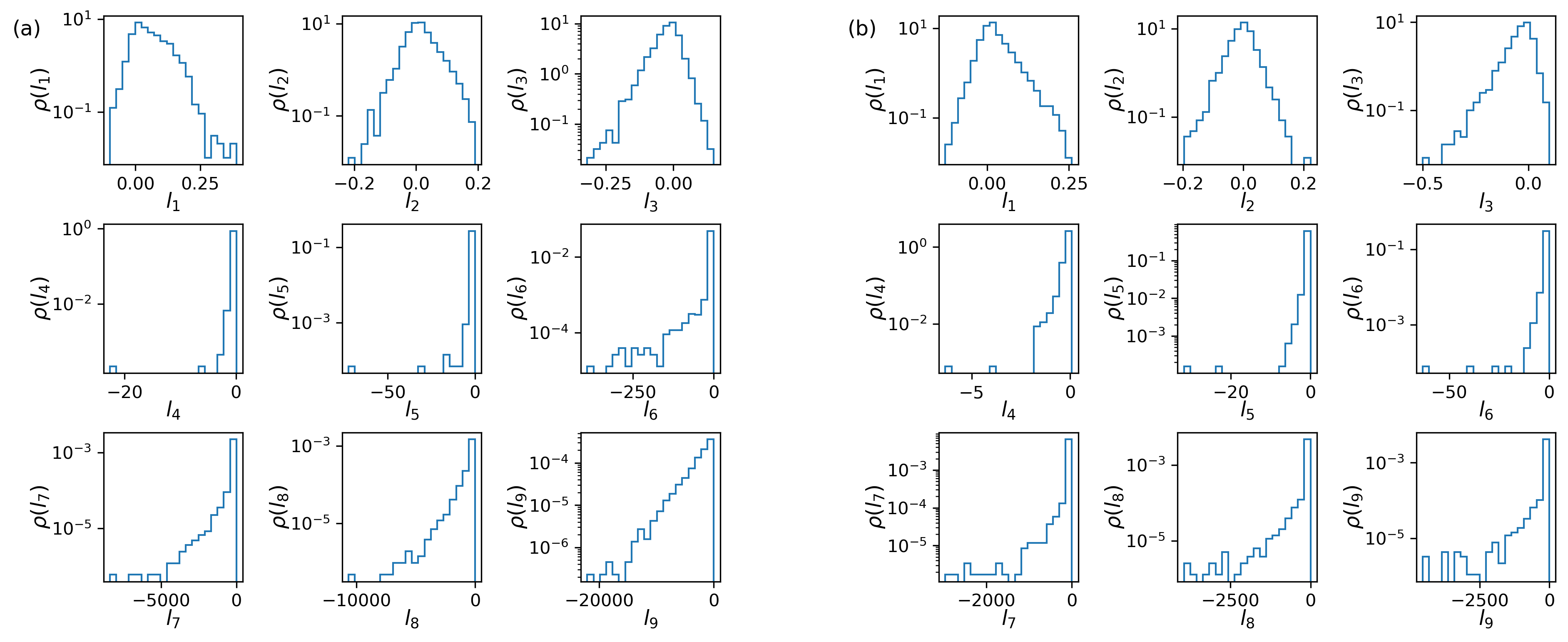}
  	\caption{Distributions of Lyapunov exponents computed for 8000 training runs of networks initizalized with wide-range initialization, amplifying input with (a) ReLU as activation function and (b) tanh as activation function.}
  	\label{wlehist}
 \end{figure*}
To characterize the dynamical properties of our sample artificial neural network we compute indicators of local stability, i.e., Lyapunov exponents (LEs) $l_q$, finite-time-Lyapunov exponents (FTLEs) $\lambda_q$ and covariant Lyapunov vectors (CLVs), with  with $q \in (1, Q)$ and $Q$ being the dimension of the system under study, i.e. here Q=9.
While LEs measure the growth rate of perturbations in different directions of the phase space, the CLVs align with the directions of perturbation growth within a dynamical system.
In this contribution LEs are computed using Benettin's algorithm \cite{benettin1978all}. We compute CLVs via Ginelli's \cite{Ginelli} method as well as an estimation method proposed in \cite{Sharafi}.
The later method is employed the purpose of predictions (Sec.\ref{pred}), since it allows us to get an estimate of the covariant Lyapunov vectors without knowing the system's trajectory in the far future.
All algorithms (Bennetin, Ginelli, Sharafi) require an integration of the respective dynamical system and the Jacobian (or an estimate of the Jacobian \cite{Martin2022estimating}) of the dynamical systems.
%
\begin{figure*}[t!]
  \includegraphics[width=\textwidth]{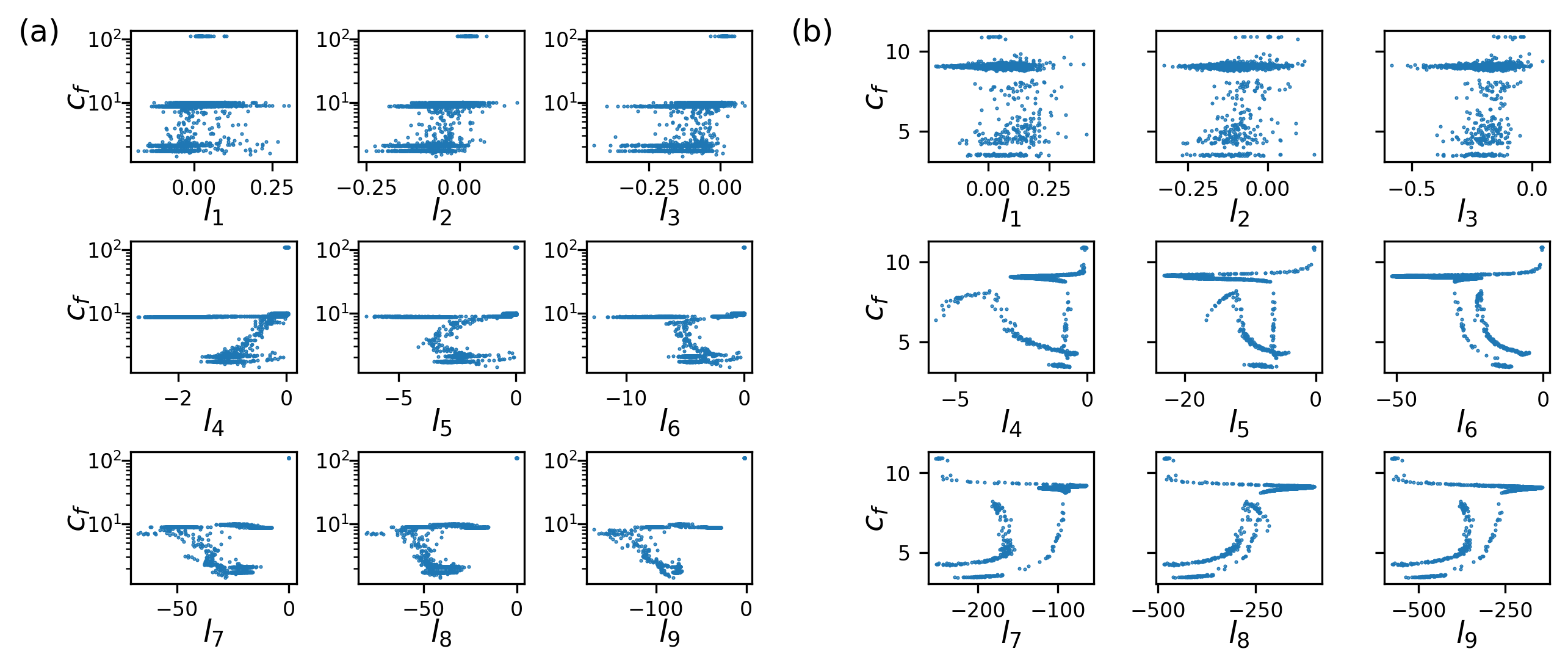}
  \vspace{-0.8cm}
 	\caption{Relations of stable LEs and final loss-values can exhibit nonlinear structures and are not random, whereas less stable LEs appear to have arbitrary values. To produce these results, He initialization has been applied. (a) $l_q$ - $c_f$-relations of 4000 network realizations using ReLU as activation. b) $l_q$-$c_f$ relations of 4000 network realizations using tanh activation.}
 	\label{2lossle}
 \end{figure*}

Using Eqn.~(\ref{w21dot}) -- (\ref{b2dot}) we can integrate the artificial neural network as a dynamical system.
The Jacobian of this dynamical system is given by Eqn.~(\ref{J1}) -- (\ref{J16}) and allows to compute FTLEs and CLVs at each point of the trajectory (i.e., during the network-training phase).
In this context, the learning rate of the training process is identical to integration step $dt$ of the dynamical system.
For all training runs under study, the learning rate (integration step) is set to $dt = 0.00003$.
As a suitable value for the orthogonalization interval required for all three algorithms we use $d\tau = 0.0006$, i.e., every 20 integration steps, the evolving perturbations are orthogonalized.
In order to capture the dynamical properties as soon as possible we use both algorithms without an initial transient.
To estimate CLVs we use a backward iteration interval of 0.006, that is 10 orthogonalization intervals.
For each network realization we have trained the network for 733333 time steps, that is 36667 orthogonalization intervals.
Working with a batch size of 32, this amount of backpropagations corresponds to a training lengths of 23467 epochs.
Although such long training runs are rare to find in practical applications, one can argue that one can expect the values of the loss-function to have converged to the closest local minimum within this long training time.
\begin{figure*}[t!]
    \includegraphics[width=\textwidth]{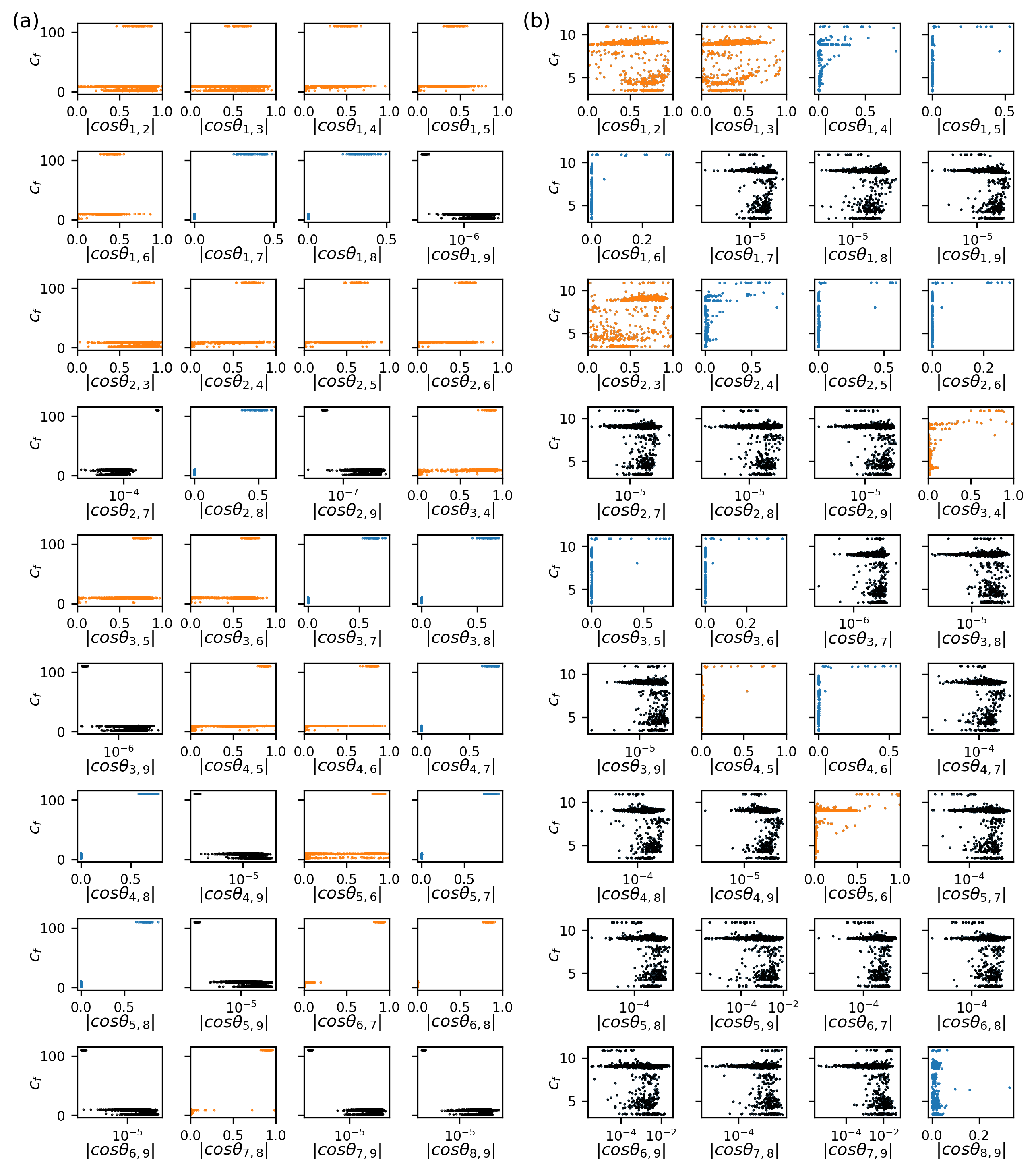}
    \vspace{-0.5cm}
    \caption{Relations of angles between CLVs at the end of training and final loss-values can exhibit nonlinear structures. To produce these results, He initialization has been applied. (a) $|\mbox{cos}\,\theta_{i,j}|$ - $c_f$-relations of 4000 network realizations using ReLU as activation. b) $|\mbox{cos}\,\theta_{i,j}|$ - $c_f$ relations of 4000 network realizations using tanh activation. Different colours emphasize the existence of tangencies between CLVs (red), i.e.,
      $|\mbox{cos} \theta_{i,j} \approx 1|$  or mostly orthogonality ($|\mbox{cos} \theta_{i,j} \approx 0$|) of CLVs (black), or intermediate behaviour (blue)}
   	\label{2losscos}
  \end{figure*}


Applying the specifications indicated above, we compute LEs for networks initialized using He initialization as well as wide range initialization. 
Lyapunov exponents (averaged over the whole training time) of 8000 different runs using He initialization are shown in Fig.~\ref{lehist}.
Comparing the results of training with ReLU activation (Fig.~\ref{lehist}(a)), to the corresponding results with tanh activation (Fig.~\ref{lehist}(b)), regions in the phase space of the network trained with tanh activation appear to have stronger contracting directions with more negative Lyapunov exponents than the network trained with ReLU.
33\% of the networks trained with ReLU activation and 62\% of the networks trained with tanh activation are in the chaotic regime (i.e. first LE is positive). 
In the case of the networks trained with ReLU activation, all other exponents except the last one can also become positive (3\% of the networks have their three leading Lyapunov exponents above zero). In the case of networks trained with tanh activation only the first two leading exponents are sometimes positive (none of the networks have three positive Lyapunov exponents).

\begin{figure*}[t!]
   \includegraphics[width=\textwidth]{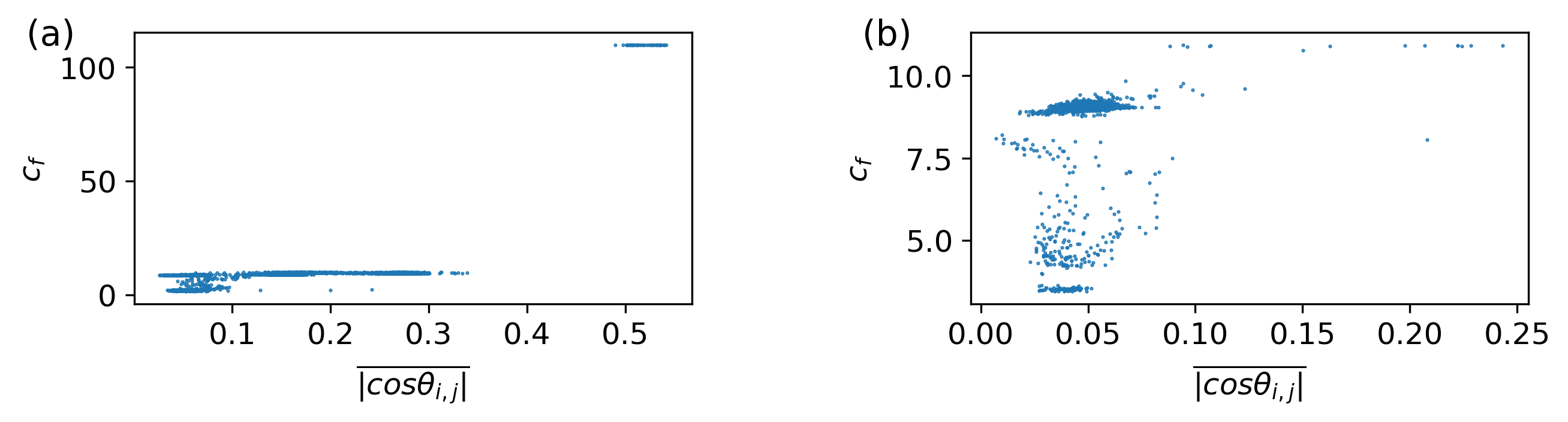}
   \vspace{-0.5cm}
 	\caption{ The clear reduction in the average angle seen between the covariant Lyapunov vectors in high final loss values suggest a phase transition. To produce these results, He initialization has been applied. (a) $\overline{|cos\theta_{i,j}|}$ - $c_f$-relations of 4000 network realizations using ReLU as activation. b) $\overline{|cos\theta_{i,j}|}$ - $c_f$ relations of 4000 network realizations using tanh activation.}
  	\label{avglosscos}
 \end{figure*}
As can be seen in Fig.~\ref{wlehist}, wide range initialization of both network types tested produced highly negative Lyapunov exponents.
However the distributions of LEs suggests that these highly negative values are rare and most of the mass of the distributions for wide-range initialized networks trained with ReLU are located around zero, supporting the idea that wide-range initialization leads to a loss in stability.
It is also interesting to note that  the distributions of the non-leading LEs tend to have more than one maximum if the networks were initialized using He initialization. This indicates that we may be able to identify different attractors using the non-leading Lyapunov exponents.
Additionally several distributions are decaying according to an almost perfect exponential law. 

In order to identify possible correlations between LEs and final losses, we also plotted LEs versus $c_f$ values for the same 8000 runs using He initialization in Fig.~\ref{2lossle}.
Note that the final losses of the networks trained with ReLU activation function are plotted on a logarithmic scale.
In the networks using ReLU for signal enhancement (Fig.~\ref{2lossle} (a)), the first three LEs show no clear association with intermediate or low loss values (clusters of points forming horizontal lines).
Whereas very high values of the loss function (upper edge in the first row of graphics in Fig.~\ref{2lossle}(a)) only occur if the first three LEs are larger than zero.
These high values of the loss function occur also simultaneously with stable LEs having values larger than or close to zero, indicating loss of stability in all directions being associated with high final loss values.
The occurrence of these high values of the loss functions is relatively rare according to Fig.~\ref{2losshist} (a), however the probability of observing these effects is clearly not zero.

For $l_4$ - $l_9$, it is not possible to formulate the $l_q$ - $c_f$-relation as simple tendencies. Nevertheless it is interesting to note nonlinear-structures in the $l_q - c_f$-relation that are clearly not random.
 For networks applying tanh activation (see Fig.~\ref{2lossle}(b)), we find qualitatively similar results for LEs with an index smaller than 3, i.e. very pronounced structures in the $l_q$- $c_f$-plane, indicating non-trivial, non-linear relations between values of the cost functions and stable LEs.
 However no conclusions can be drawn about an association of loss values and ranges of LEs for the three leading exponents.
 Note that when training the network with tanh activation very high final loss values were also not observed at all, i.e. the high values of $c_f$ for tanh activations are in the range of intermediate loss values for ReLU activation.
As can be seen from the distributions of the Lyapunov exponents in  Figs.~\ref{lehist}, it is highly unlikely for exponents $l_4$ - $l_6$ in the He initialization  and tanh training scheme to get close to zero, while $l_7$-$l_9$ are always highly negative. Thus the phenomenon of loss of stability in all directions is not observed.   

 The corresponding figures for networks initialized with wide-range initialization show -- in contrast to the findings for He-initialization -- structures that appear to be almost random and are therefore presented and discussed in the appendix.

In addition, we  visualize relations between final loss values and angles between CLVs in Fig.~\ref{2losscos}.
One can basically distinguish three different types of behaviour: vectors that can become (approximately) tangent to each other (cos$\theta_{i,j} \approx 1$), vectors that are mostly orthogonal to each other (cos$\theta_{i,j} \approx 0$) and vectors that assume angles in between both.
The choice of colours in Fig.~\ref{2losscos} reflects these three groups.
Interestingly some of the vector pairs showing tangencies ((1,2),(1,3), (2,3),(3,4),(4,5),(5,6)) do so for both types of activation functions, indicating that this behaviour could probably be attributed to the structure of the network.
Similarly, the 9th vector is also mostly orthogonal to all other vectors for both types of activation functions.
Keeping in mind that the ordering of the vectors occurs according to their instability, one can in general say that tangencies are observed between the least stable vectors, whereas more stable vectors tend to be orthogonal to each other.
%
In the case of networks trained with ReLU in which high-loss values occur, the tangencies between the CLVs are associated with maximum loss values.
Tangencies between covariant Lyapunov vectors are a sign of critical transitions and a manifestation of loss of transverse stability (see  \cite{Sharafi}).
This implies that while the trajectory of the majority of the realizations stay in the same region of the phase space some realization manage to enter regions of the phase space that is accompanied by loss of transverse stability. As a result of entering these  "hot spots" any perturbation  may lead to their  transition to another attractor, which makes it impossible for them to ever reach low loss values. The different badges of input data can be interpreted in  this sense as perturbations that may facilitate a transition. 
This is in agreement with results from  Fig.~\ref{2lossle}(a) in which you can see that the occurrence of high loss value requires at least  three leading LEs to be positive and LEs corresponding to the contracting directions to be close to zero.

There are also vector pairs which display different behaviours depending on the chosen activation function.
Considering the relation between angles and values and final losses, one can also distinguish angles between CLVs that are clearly related to high values or low values of the final loss:
For networks using RELU as activation function (left side of Fig.~\ref{2losscos}), all vector pairs involving the 7th and 8th vector (except (7,9) and (8,9)) show orthogonality if low values of $c_f$ are observed and angles that clearly deviating from orthogonality if high values of the loss-function are observed.
The respective figures display two linearly separable regions, indicating that monitoring these angles would potentially allow  prediction of  values of $c_f$, as we will discuss in more detail in Sec.~\ref{pred}. 
If tanh is applied as activation function, very high values of $c_f$ , i.e., $c_f \approx 100$) are not observed (in comparison to networks with RELU).
Nevertheless the angles between vector pairs (1,5), (1,6), (2,5), (2,6), (3,5), (3,6), (4,5) and (4,6) also display deviations from orthogonality if the loss function did not reach its optimum.

To clarify the degree of the overall tangency or orthogonality of the CLVs at the end of the training we show the sum  of the absolute value of the cosine of the angles between all the vectors divided by the number of the angles ($\overline{|cos\theta_{i,j}|}$) at the end of training  versus the final loss in Figs.~\ref{avglosscos}. 
The average angle between the vectors for networks trained with ReLU activation function shows a very distinct decrease for trainings that have a high final loss value (Figs.~\ref{avglosscos}.a, upper right corner).
In this sense the high loss values are not only statistically but also dynamically different from the low loss results. 
The qualitative change in the dynamical properties suggest that we may be able to use dynamical indicators, albeit in the beginning of the training, to predict high loss values effectively.

The lowest loss values do not possess such clear change in the dynamical properties, however they are  more often than not associated with a high degree of orthogonality between the vectors. 
In the networks trained with tanh activation function (Figs.~\ref{avglosscos}.b) the very high loss values observed  with ReLU activation are absent. There is not a prominent decrease in the angle between the vectors, which implies that no transition occurs in the dynamics of the realizations with Tanh activation function and He initialization. However the same as with the ReLU activation function  the lowest loss values tend to have higher degree of orthogonality between the vectors. 
%
\section{Predicting outcomes of training processes by monitoring local stability}
%
\begin{figure*}[t!!!]
 	\centering
 	\includegraphics[width=\textwidth]{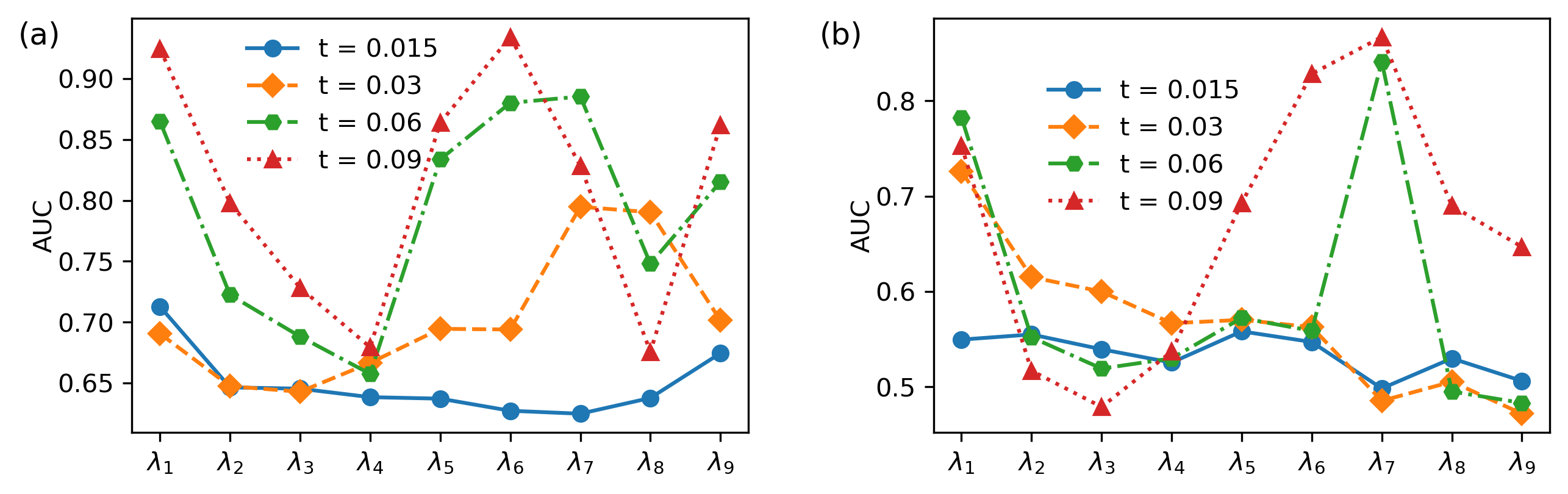}
        \vspace{-0.8cm}
 	\caption{ AUCs for predicting low final losses with FTLEs ($\lambda_i$) at different times during training show that the FTLEs can predict a low final loss (values below 4) as early as $t = 0.03$ into the training. He initialization method has been applied. (a) AUCs using 34000 network realizations using ReLU as activation. b) AUCs using 34000 network realization using tanh activation.}
 	\label{2auclexbelow3and4}
 \end{figure*}
\label{pred}
Finding some non-trivial relations of LEs as well as the CLVs  and final loss values, one can ask, whether these observations are frequent enough to justify saying that training outcomes are linked to local stability of the underlying dynamical systems.
In order to determine the existence  of these links, we follow the idea of Granger causality \cite{Granger, Kim2011} and test whether it is possible to predict ideal training runs (very low values of the final loss $c_f$ or failed training runs (very high values of $c_f$) by monitoring FTLEs and CLVs during the training process..
%
In more detail, we test if the values of FTLEs and CLVs can indicate, early on in the training, if the network will convert to a specific region in the phase space. 
In the following we focus on networks initialized with He initialization and divide the outcomes of 8000 training runs into training and test data sets of equal length.
To apply CLVs in a predictive context, we estimate them at every time step of each training run using the method introduced in \cite{Sharafi}, i.e., without the knowledge of the far future of the trajectory.
We then train a naive Bayesian classifier on the training data-set and predict the occurrence of very high as well as very low losses given the FTLEs and the CLVs.
A naive Bayesian classifier is chosen, since it does not depend on any other parameter choice than the number of bins used to estimate predictive Distributions (conditional probability distributions).

More precisely, we aim to predict at time steps $t = 0.015$ corresponding to 500 orthogonalization steps (320 epochs),  $t = 0.003$ corresponding to 1000 orthogonalization steps (640 epochs), $t = 0.06$ corresponding to 2000 orthogonalization steps (1280 epochs) and $t = 0.009$, corresponding to 3000 orthogonalization steps (1920 epochs), whether or not $c_f$ (measured after 23467 epochs) is below 4 (low loss case) or higher than 100 (high loss case, only for networks trained with ReLU activation function).
We analyze the results by calculating contingency tables, Receiver Operating Characteristic curves (ROC curves) and the area under ROC curves (AUCs) as a measure of prediction success.
 \begin{figure*}[t!]
 	\centering
 	\includegraphics[width=\textwidth]{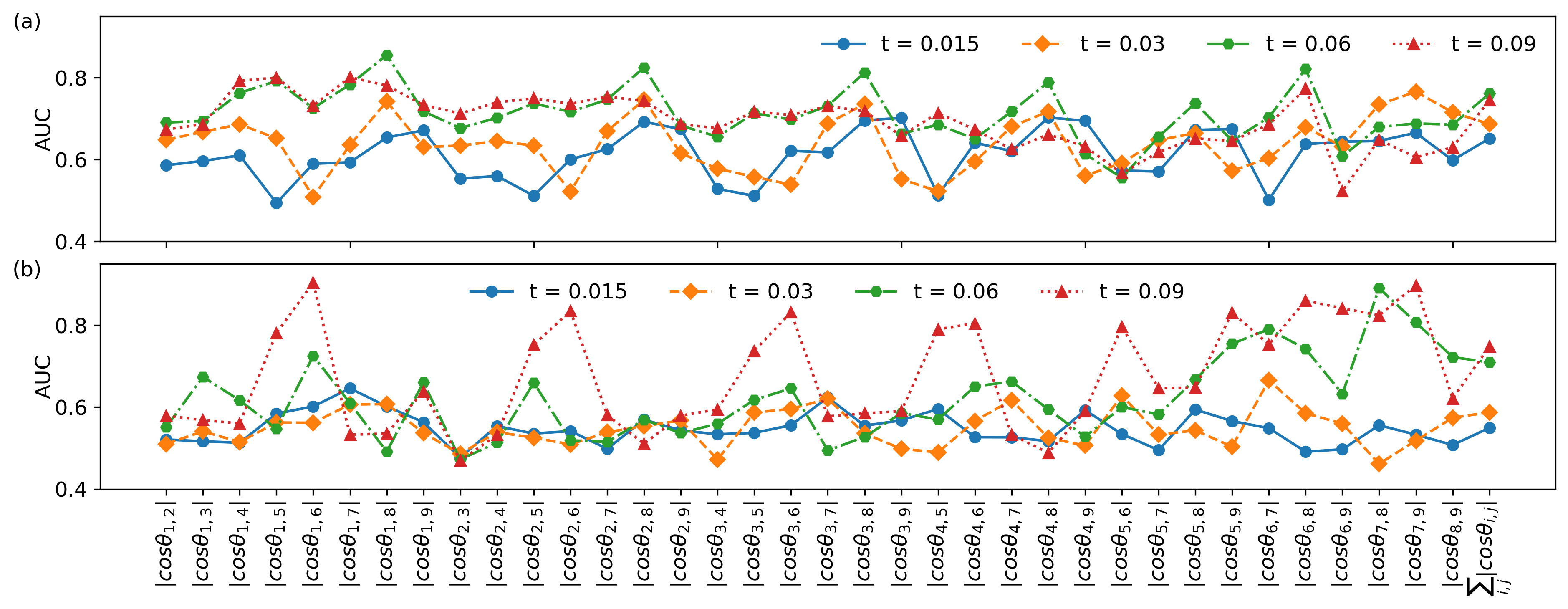}
        \vspace{-0.8cm}
 	\caption{AUCs of low final loss values (below 4) show that the cosines of the angles between some of the CLVs ($\theta_{i,j}$) can be used as predictors.  (a) AUCs using 8000 network realizations using ReLU as activation. b) AUCs using 8000 network realization tanh activation.}
 	\label{2auccosbelow3and4}
 \end{figure*}
 %
AUC values of unity indicate an optimal classifier, whereas AUC values of $0.5$ characterize randomly made binary classification.
The values of the AUCs for different FTLEs and at different times during the training process, predicting low final loss values (below 4) are shown in Fig.~\ref{2auclexbelow3and4}.
While most values are above 0.5 and therefore better than random predictors, the first the 6th and the 7th Lyapunov exponents well if either ReLU or Tanh activation functions are in the network. 
Additionally, we predict ideal training runs (low final loss) given the angles between CLVs (see Fig.~\ref{2auccosbelow3and4}).
While the cosines of angle between some CLVs are not significant as precursors others can effectively predict the final %
loss of the trained network. 

\begin{figure}[t!!!]
 	\centering
 	\includegraphics[width=0.45\textwidth]{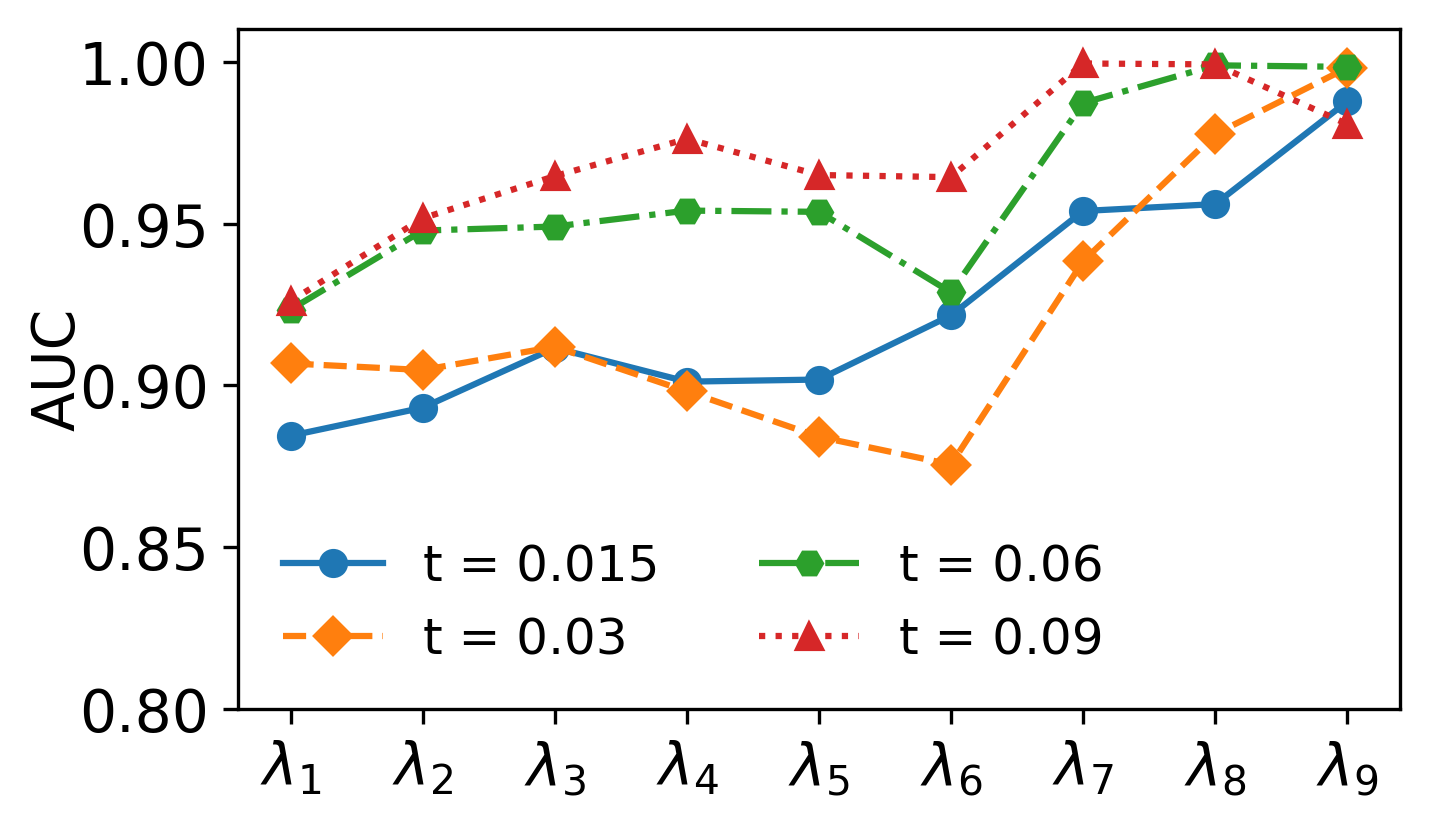}
        \vspace{-0.5cm}
 	\caption{ AUCs for predicting final high losses (values above 100) with FTLEs ($\lambda_i$) at different times during training indicate FTLEs are effective early predictors of high loss values. 8000 network realizations using ReLU as activation. He initialization method has been applied. }
 	\label{auclexabove100}
 \end{figure}

Since training the network with ReLU activation can lead to high values of the final loss function we also tested for the predictive powers of FTLEs and CLVs in the high loss limit (values above 100) while training with ReLU activation (see e.g. Fig.~\ref{auclexabove100}).
%

 \begin{figure*}[t!!!]
 	\centering
 	\includegraphics[width=\textwidth]{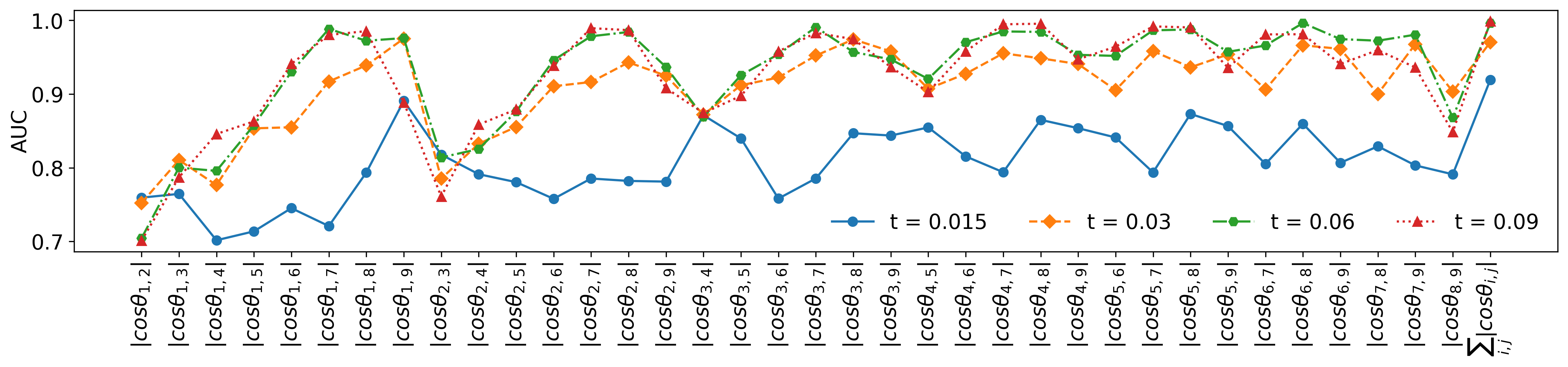}
        \vspace{-1cm}
 	\caption{ AUCs of high final loss values show that the  cosines of the angles between the CLVs ($\theta_{i,j}$) can predict high loss values (above 100) effectively. 8000 network realizations using ReLU as activation.}
 	\label{auccosabove100}
 \end{figure*}
In this limit all FTLEs are quite effective predictors.
Specifically the three most stable exponents can predict the final loss value being above 100 with high accuracy even as early as $t = 0.015$.
 The results for predicting high loss values for a network trained with ReLU activation using the angels between the CLVs are presented in Fig.~\ref{auccosabove100}. The sum of the absolute value of the cosines of the angles between vectors seem to be the most effective predictor.

 High loss prediction is relevant for practical purposes, since the training run could be stopped and the network reinitialized, if it becomes very likely that the process is converging to a suboptimal minimum of the cost-function.
This is less relevant for a small network as it was studied here which can be trained quickly, but more relevant for monitoring the training of larger networks.
Related ideas (so called {\sl early stopping}) for ending the training process (but without re-initialization and restart) are based on monitoring the generalization error (difference of loss-function for test and training data set).
In order to quantify the relevance of FTLEs for practical purposes, we compare the best predictions made using the FTLEs, angles between the CLVs and predictions based on time-series of the cost-function in table \ref{auctable}.
In more detail, we focus on 8000 networks using RELU for signal enhancement and choose the best predicting angle between the CLVs as well as the FTLEs in addition  to the values of the loss function $C(t)$ at the respective time steps for predicting high training-loss values using a Bayesian classifier.
These results are compared to the predictions of all nine $FTLE$-values combined through a support vector machine (SVM) classifier with a radial-basis-function kernel. 
Although the results using FTLEs combined through a SVM-classifier are slightly better, than results obtained through monitoring the loss-function, it is questionable, whether or not the numerical effort to estimate FTLEs justifies this improvement in predictability.
%
\begin{table*}[t!]
  \label{auctable}
  \caption{AUC for predicting high loss ($c_f > 100$) using various predictors. The respective 8000 networks are initialized with the He method and signal enhancement is using RELU.}
    \begin{ruledtabular}
  \begin{tabular}{c d d d d} 
  $t$ & 0.015 & 0.03 & 0.06 & 0.09 \\ [0.5ex] 
 \hline\hline
 $\mbox{cos}\,\theta_{i,j}(t)$ & 0.919201 & 0.975243 & 0.996333 & 0.99861\\
 \hline
 $\lambda_i(t)$ & 0.987756 & 0.998141 & 0.998909 & 0.999476 \\
 \hline
 SVM & 0.951719 & 0.997852 & 0.998736 & 0.998989\\
 \hline
 $C(t)$ & 0.960493 & 0.97232 & 0.97232 & 0.98918\\ 
\end{tabular}
\end{ruledtabular}
\end{table*}
It would be nevertheless interesting to test in future experiments, whether FTLEs can provide more significant improvements for the training of larger networks.

 In summary, it is possible to identify some of the many indicator variables for local stability tested (9 FTLEs and 36 angles between CLVs) that can predict low values of $c_f$ with an AUC clearly larger than 0.5.
 In particular the (marginally) stable directions (5th ,6th, 7th and 8th CLV) and their corresponding growth rates (FTLEs) seem to be relevant for these ideal training runs.
Nevertheless, there also exist FTLEs and angles between CLVs that don't display a significant relation to low final loss values. 
In contrast to this, all indicator variables are able to predict very high values of the final loss (failed training runs), as observed if signal enhancement is done using ReLU activation.
  In both test cases (predicting low and high loss values) FTLEs are capable of predicting the final loss earlier than angles between CLVs.
  Since FTLEs are also easier to estimate, monitoring FTLEs is probably more relevant for practical purposes than estimating angles between CLVs.
 In total, this suggests the existence of strong links between local stability of the training process and learning success which are worth being explored further for other network architectures and learning tasks. 
	\section{Conclusions}
	\label{con}
	\begin{figure*}[t!!!]
   \includegraphics[width=\textwidth]{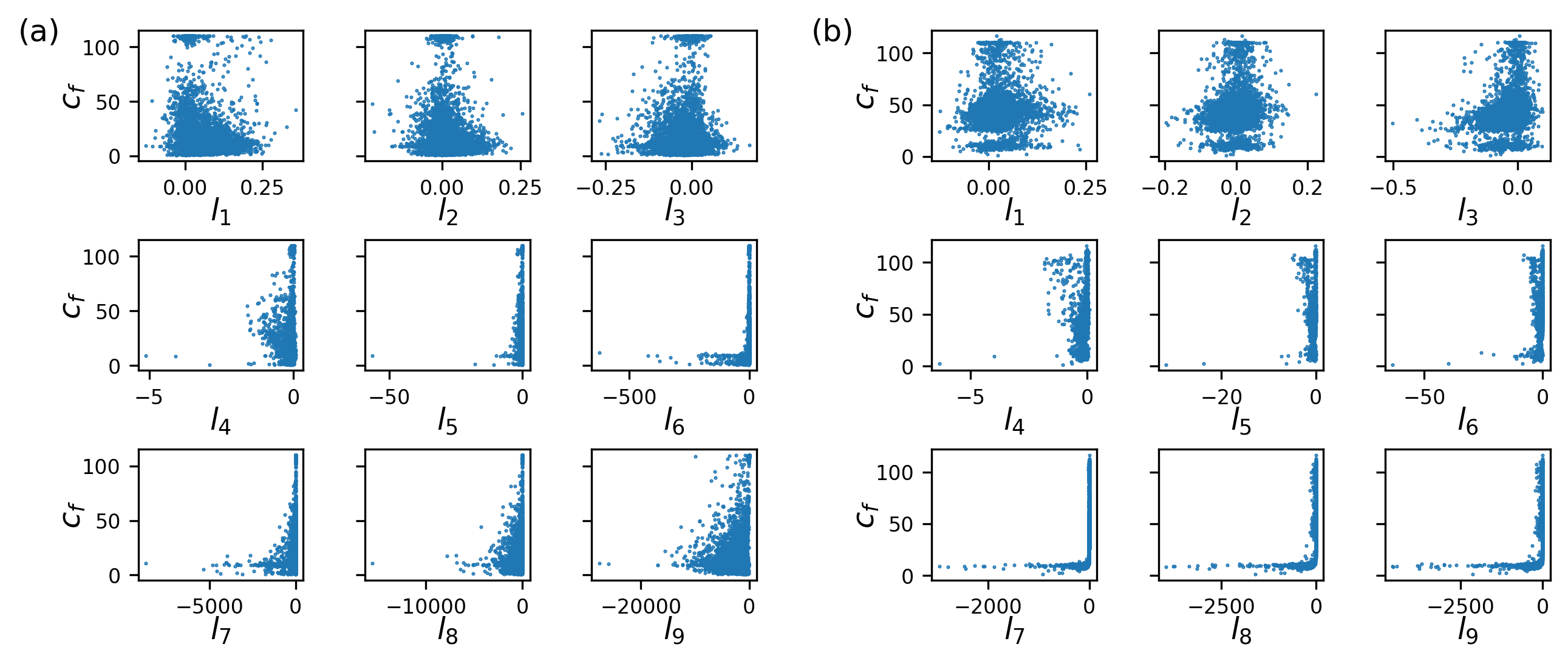}
   \vspace{-0.8cm}
 	\caption{ If wide range initialization is applied, the relation of final losses and LEs have less distinct patterns (compared to He initialization). (a) $l_q$-$c_f$-relations of 4000 network realizations using ReLU as activation. (b) $l_q$-$c_f$-relations of 4000 network realizations using tanh activation.}
 	\label{2wlossle}
 \end{figure*}
\begin{figure}[h!]
  \centering
  \includegraphics[width=0.35\textwidth]{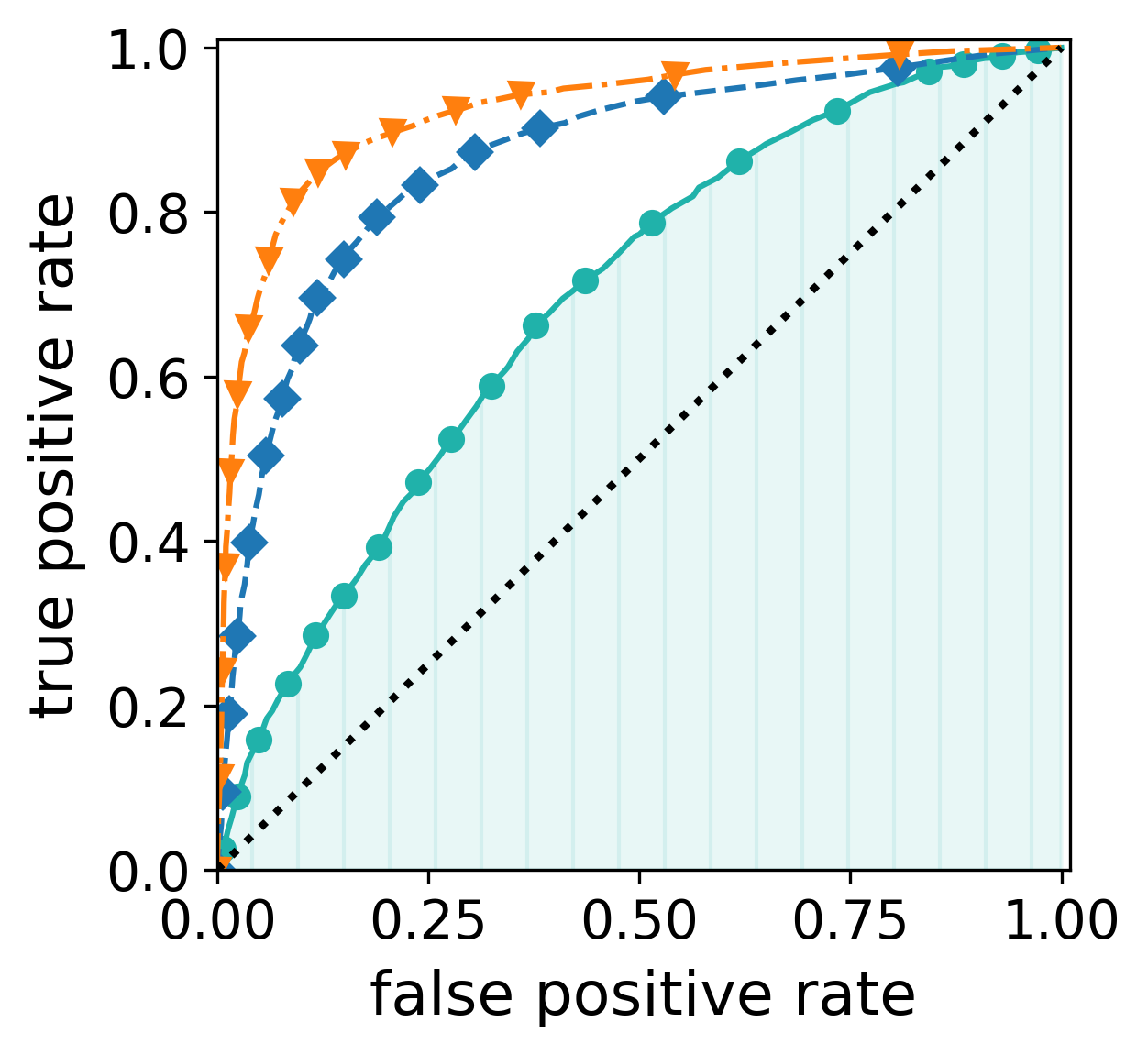}
 \caption{\label{roc} The area under an ROC curve (AUC) is a measure of the success of a precursor.}
\end{figure}
	In this contribution, we describe the weight dynamics during the training-process of feed-forward-networks with an arbitrary choice of activation function as a dynamical system.
	More specifically, we derive the equations of the learning dynamics for weights in 3-layer-networks and the equations of the Jacobian for %
	the corresponding dynamical system.
	%
   While the computational results we present here are for a specific regression task solved by a small network, the formulas %
    represented in Sec.\ref{jac} can be applied to any feed-forward network with 3 layers, arbitrary number of nodes and mean-squared-error cost function.

	Computing the Jacobian of the learning dynamics allows us to undertake a detailed analysis of the stability properties of learning 3-layer networks.
	Although the training process of a network is transient dynamics, we estimate indicators of local stability that are typically applied to systems that have converged towards an attractor.
        We identify relations between indicators of local stability and the final outcome of the training process.
	As expected the nature of the learning dynamics can depend on the learning task, the choice of activation functions, the network structure and the initialization of weights.
	The latter aspect appears to be obvious, if the system is interpreted as a non-linear dynamical system, but less obvious from a machine-learning perspective.
	Moreover, allowing for atypical choices of weight initializations one can uncover a surprisingly rich variety of learning outcomes. 
	This (unwanted in applications) variety of learning dynamics also explain the success of commonly recommended methods of network initializations \cite{he2015delving, glorot2010understanding, kumar2017weight}, since they typically prevent these rich variety of learning dynamics to appear and favour runs that converge faster to low minima.
	Our findings provide an additional understanding for this success by investigating the distributions of final loss values and the corresponding stability properties.

	One long-known short-coming of training of feed-forward-networks through gradient-descent methods is that can get stuck in one of many possible sub-optimal local minima of the loss-function \cite{gori1992problem}, \cite{auer1995exponentially}.
	Several methods and practices have been proposed in order to prevent this from happening \cite{zuge2023weight}.    
	Describing the learning network as a dynamical system, the question of whether or not a training-run can reach a low-minimum of a loss-function, depends on the initialization of network-weights.

   We demonstrate that even with common choices for initialization (He method) of the weights, training with a ReLU activation function can lead to very high final loss values. Measuring the first Lyapunov exponent, we show that a high final loss value is more likely to occur if the dynamical system is in a chaotic regime. 
	More specifically, we find that the high final loss values occur only when the three leading Lyapunov exponents are positive while the other exponents are also close to zero or slightly positive. The increase in the values of the growth rate of orthogonal directions indicates loss of transverse stability. 
	We observe that the loss of stability  manifests itself in the tangencies between the covariant Lyapunov vectors. We argue that the loss of transverse stability accompanied by  tangencies between the CLVs indicate the trajectory entering a region in the phase space where any perturbation may result transitioning to another attractor. The perturbations are provided by the changes in the input of the network. 
	
	We demonstrate that the high loss values are not only statistically but also dynamically different from the lower loss values. Therefore, we can use indicators of dynamical stability to predict the occurrence of high loss values.

	In this contribution we discussed results for one specific network structure, two common choices of activation functions and regression as as learning tasks. 
	We are aware, that analogous investigations on different learning tasks, different activation functions and different network structures could reveal qualitatively different results and we hope that this study will inspire such investigations in the future. 
	%
\section*{Acknowledgments}
The authors of this study are grateful to the BMBF for financial support within the project DADLN (01$\mid$S19079) and to the Landes\-forschungs\-f{\"o}rderung Hamburg for financial support within the project LD-SODA (LFF-FV90).

\section{Appendix}

\subsection{Relations of LEs and $\mathbf{c_f}$ for Wide-Range Initialization}
In Fig.~\ref{2wlossle} we present values of LEs versus the final loss values for network realizations initialized with wide range distribution of initial weight values.
For the leading three exponents the relation of $l_q$ and $c_f$ appear to be almost random.
For $l_4$ -- $l_6$ the range of possible values is mostly restricted to negative values close to zero, with a few exceptions of lower values occurring for lower losses.
The same is observed for very stable exponents $l_7$ -- $l_9$ in the case of tanh activation, with more pronounced low values being observed for low loss values.
If ReLU is chosen as activation function, the observed range of negative $l_7$ -- $l_9$ increases, the smaller the observed values of $c_f$ are.

\subsection{Example for an ROC-curve}
For illustrative purposes Fig.~\ref{roc} displays one example for the ROC-curves summarized by the area-under-curve index in Sec.~\ref{pred}. The figure represents ROC curves for prediction of low loss values at different times for 340000 realizations of the network trained with ReLU activation. The three different curves represent three different times during the training that the precursors were measured ($t = 0.03, 0.06, 0.09$). 

\bibliography{sharafhaller}

\end{document}